\documentclass[10pt,twocolumn,letterpaper]{article}

\usepackage[pagenumbers]{cvpr} 

\definecolor{cvprblue}{rgb}{0.21,0.49,0.74}
\usepackage[pagebackref,breaklinks,colorlinks,allcolors=cvprblue]{hyperref}
\usepackage{bm}
\usepackage{color}
\usepackage{multirow}
\usepackage{appendix}
\usepackage{float}
\usepackage{tabularx}
\usepackage{algorithm} 
\usepackage{algorithmic}

\def\paperID{1634} 
\def\confName{CVPR}
\def\confYear{2026}

\title{Adaptive Begin-of-Video Tokens for\\Autoregressive Video Diffusion Models}

\newcommand*\samethanks[1][\value{footnote}]{\footnotemark[#1]}
\vspace{-2ex}
\author{
Tianle Cheng\textsuperscript{\normalfont 1}\thanks{Equal contribution.} \quad
Zeyan Zhang\textsuperscript{\normalfont 1}\samethanks \quad
Kaifeng Gao\textsuperscript{\normalfont 1,2}\thanks{Corresponding author.} \quad
Jun Xiao\textsuperscript{\normalfont 1}\\
\textsuperscript{\normalfont 1}Zhejiang University \quad
\textsuperscript{\normalfont 2}Manycore Tech Inc. \\
{\tt\small \{jackcheng, 22321072, kite\_phone\}@zju.edu.cn, junx@cs.zju.edu.cn}
\vspace{-.3em}
}

\begin{document}
\maketitle
\begin{abstract}
Recent advancements in diffusion-based video generation have produced impressive and high-fidelity short videos.
To extend these successes to generate coherent long videos, most video diffusion models (VDMs) generate videos in an autoregressive manner, \ie, generating subsequent frames conditioned on previous ones. 
There are generally two primary paradigms: chunk-based extension and stream denoising. The former directly concatenates previous clean frames as conditioning, suffering from denoising latency and error accumulation. The latter maintains the denoising sequence with monotonically increasing noise levels. In each denoising iteration, one clean frame is produced while a new pure noise is simultaneously appended, enabling live-stream sampling. However, it struggles with fragile consistency and poor motion dynamics.
In this paper, we propose \textbf{Adaptive Begin-of-Video Tokens} (ada-BOV) for autoregressive VDMs. The BOV tokens are special learnable embeddings on VDMs. They adaptively absorb denoised preceding frames via an adaptive-layer-norm-like modulation. This design preserves the global consistency while allowing for flexible conditioning in dynamic scenarios. 
To ensure the quality of local dynamics essential in modulating BOV tokens, we further propose a refinement strategy for stream denoising. It decouples the sampling trajectory length from the attention window size constraint, leading to improved local guidance and overall imaging quality. We also propose a disturbance-augmented training noise schedule, which balances the convergence speed with model robustness for the stream denoising. Extensive experiments demonstrate that our method achieves compelling qualitative and quantitative results across multiple metrics.
\end{abstract}

\section{Introduction}
\label{sec:intro}
\par With the significant success of diffusion models~\cite{sohl2015deep, ho2020denoising, song2021scorebased, ramesh2022hierarchical, rombach2022high}, researchers have recently demonstrated remarkable progress in generating high-quality videos with video diffusion models (VDMs)~\cite{zhou2022magicvideo, blattmann2023align, guo2024animatediff, wang2025lavie}.

Generating coherent long videos remains a challenging yet important goal. Due to the prohibitive computational cost of self-attention over extended sequences, the attention window size in existing models is typically limited. Consequently, many approaches rely on autoregressive extension techniques to generate longer videos. A typical paradigm is chunk-based extension~\cite{ho2022video, tan2024video, gao2025ca2}: They iteratively generate video chunks conditioned on previously synthesized frames, \ie, replacing portions of the noisy frame sequence with clean frames as reference.
The remaining noisy frames are denoised synchronously, sharing a unified noise level. Nevertheless, a common issue in chunk-based methods is the stark noise-level discrepancy between clean (reference) frames and noisy frames. It often leads to temporal inconsistencies and motion discontinuities~\cite{xie2025progressive, sun2025ar}.
\begin{figure*}[t]
\centering
\includegraphics[width=1.\textwidth]{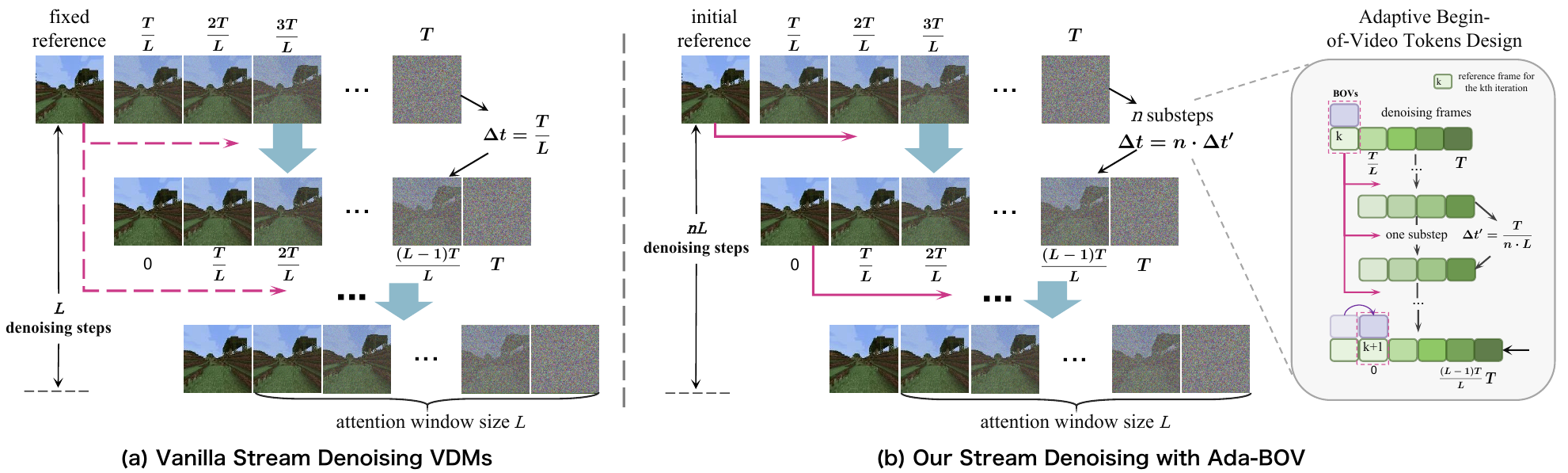}
\vspace{-3ex}
\caption{Pipeline for any frame from pure noise to clean. (a) Vanilla stream denoising VDMs. The dashed arrows indicate that a fixed global reference frame can be optionally injected~\cite{henschel2025streamingt2v, tian2024videotetris}. (b) Our approach, where the solid arrows indicate the injection of reference frames. The BOV tokens contain a variety of global features and are modulated with the latest generated frame for flexible guidance. And the refinement strategy for stream denoising supports denser denoising steps.}
\label{fig1}
\end{figure*}

\par To address this issue, more recent works~\cite{kim2024fifo, ruhe2024rolling, xie2025progressive} propose to assign monotonically increasing noise levels to the entire denoising sequence. 
As illustrated in Fig.~\ref{fig1}(a), in each iteration, one clean frame is output and a new pure-noise frame is appended, \ie, forming a \textbf{stream denoising paradigm}. It not only mitigates the visual artifacts but also reduces denoising latency, enabling live-stream generation.
Some methods~\cite{long2024videostudio, tian2024videotetris, zeng2024make} further leverage additional global guidance throughout the generation process to mitigate error accumulation, enhancing both subject consistency and overall frame coherence.

\par However, despite these advances, we argue that current methods still suffer from several fundamental drawbacks:
1) \textbf{Fragile global consistency.} Injecting fixed global guidance is a common yet brittle strategy. While effective for maintaining subject identity in static scenes, its relevance rapidly diminishes in dynamic scenarios (\eg, scene transitions and complex camera movements). This constraint from fixed reference frames consequently hinders the model's flexibility and adaptability.
2) \textbf{Compromised motion dynamics.} 
On the other hand, discarding the additional clean frames makes the model rely solely on noisy frames, which serve as references for each other.
We argue that this can lead to a suppression of inter-frame motion, ultimately compromising the motion dynamics of the generated videos. As evidenced in prior work~\cite{xie2025progressive}, there is a strong correlation between the noise level of reference frames and motion dynamics. With clean references, the chunk-based extension excels in motion dynamics. In contrast, with noisy references, its performance collapses and falls behind the stream denoising paradigm.
3) \textbf{Limited Sampling Trajectory Length.}
For vanilla stream denoising, the sampling trajectory~\cite{song2021denoising} length (\ie, the number of denoising steps) is inherently tied to the attention window size\footnote{In this paper, we refer to the sequence to be denoised in each model forward pass as the attention window.}: A short attention window means a small number of sampling steps, resulting in low-quality outcomes. Increasing the attention window size for longer sampling trajectories leads to a drastic increase in computational overhead.

\par Based on such considerations, in this paper, we propose \textbf{adaptive Begin-of-Video tokens} (ada-BOV) for autoregressive VDMs.
This design addresses the rigidity of fixed global references~\cite{tian2024videotetris, weng2024art, zhao2025moviedreamer} and the inherent ambiguity of relying solely on noisy frames~\cite{kim2024fifo, xie2025progressive}.
A set of learnable embeddings (BOV tokens) is trained to implicitly capture the essential global features required during inference. 
Meanwhile, to achieve fine-grained temporal control, periodically updated clean references are used to modulate the BOV tokens dynamically. This design addresses the rigidity of fixed references and also significantly enhances motion dynamics by providing higher-fidelity guidance.



\par To ensure high-quality local dynamics for modulating BOV tokens, we further propose a refinement strategy for the stream denoising paradigm. It decouples the sampling trajectory length from the attention window size by executing multiple sub-denoising steps within each iteration. This effectively increases the sampling trajectory, improving both local guidance and overall imaging quality. 
Meanwhile, we investigate various training noise schedules and propose a disturbance-augmented strategy for stream-denoising model training. It balances the training convergence speed and model robustness, offering valuable insights for other stream denoising VDMs.


\par We conducted extensive evaluation experiments on two public datasets: Minecraft~\cite{guss2019minerl} and Sky Timelapse~\cite{zhang2020dtvnet}. The results demonstrate that our model significantly improves both visual quality and temporal dynamics over existing baselines.
\par In summary, we make the following three contributions:
\begin{itemize}
    \item An adaptive BOV design for autoregressive VDMs. It not only combines global features and local dynamics but also provides high-fidelity guidance for the stream denoising paradigm.
    \item A refinement strategy for stream denoising consists of a novel inference strategy and a disturbance-augmented training noise schedule. The former decouples sampling trajectory length from attention window size, and the latter balances convergence and robustness.
    \item Extensive experiments demonstrate that our method achieves state-of-the-art performance on long video generation, all while maintaining computational efficiency. 
\end{itemize}

\section{Related Works}
\label{sec:related}
\noindent\textbf{Video Diffusion Models.}
Diffusion models have significantly advanced the field of video generation. Early influential works~\cite{singer2023makeavideo, blattmann2023stable, chen2024videocrafter2} extend the 2D U-Net architecture~\cite{ronneberger2015u} into a 3D structure~\cite{cciccek20163d} capable of capturing spatio-temporal dependencies. More recently, following the success of diffusion Transformers (DiT)~\cite{peebles2023scalable} in image generation, a new trend of adopting Transformer-based backbones has emerged and yielded promising results~\cite{kong2024hunyuanvideo, yang2025cogvideox, hacohen2025ltx, wan2025, opensora2}.

\vspace{4pt}
\noindent\textbf{Chunk-Based Extension.}
As the most common and straightforward approach for long video generation, it iteratively generates a new clip conditioned on previously synthesized frames~\cite{ho2022video, he2022latent, chen2023seine, opensora}.
However, such an iterative process is prone to error accumulation, often leading to temporal drift and a degradation of long-term global consistency. To mitigate this issue, many methods inject global feature guidance additionally, such as via cross-attention~\cite{zhang2023i2vgen, ouyang2024flexifilm, tian2024videotetris}, channel-wise concatenation~\cite{zeng2024make, chen2024livephoto, xi2025omnivdiff}, or temporal-wise concatenation~\cite{esser2023structure, xing2024dynamicrafter}. A common strategy is to select a fixed frame as a global reference. For instance, ART$\cdot$V~\cite{weng2024art} leverages the first frame to preserve identity. However, this reliance on fixed references critically limits adaptability in dynamic scenarios with significant content shifts or scene transitions.

\vspace{4pt}
\noindent\textbf{Stream Denoising.}
Inspired by Diffusion Forcing~\cite{chen2024diffusion}, the stream denoising paradigm~\cite{kim2024fifo, ruhe2024rolling, xie2025progressive, sun2025ar} provides an alternative autoregressive approach. Instead of conditioning on fully denoised frames, it applies independent, monotonically increasing noise levels to each condition frame during inference. While this method enhances local coherence and accelerates inference, it introduces distinct drawbacks. Conditioning solely on noisy frames can suppress motion dynamics, and the generation quality is often sensitive to the attention window size. In contrast to existing works, our ada-BOV token design enhances both global consistency and motion dynamics, and the refinement strategy for stream denoising decouples model performance from the constraints of the attention window size.
\section{Method}
Our method is illustrated in Fig.~\ref{fig1}(b).
We first introduce the adaptive Begin-of-Video (ada-BOV) token design in Sec.~\ref{sec:bot}. Then, we elaborate on the refinement strategy for stream denoising in Sec.~\ref{sec:RSD}. 
In this section, $L$ represents the attention window size, and $i$ represents the frame index. The maximum diffusion timestep is given as $T$, with each timestep denoted as $t=1,2,\ldots, T$. 

\subsection{Adaptive Begin-of-Video Tokens}\label{sec:bot}
\noindent\textbf{Preliminaries.} Diffusion Models~\cite{sohl2015deep, ho2020denoising} aim to approximate a data distribution $q(\bm{z})$ by defining a diffusion process and learning a reverse (denoising) process. In the diffusion process, each diffusion step for frame $\bm{z}$ is $q(\bm{z}^{t}|\bm{z}^{t-1}) = \mathcal{N}(\bm{z}^t;\sqrt{\alpha^t}\bm{z}^{t-1},(1-\alpha^t)\bm{I})$,
where $\alpha^t \in(0,1)$ is the variance schedule. By applying the reparameterization trick~\cite{ho2020denoising}, it can be expressed as $\bm{z}^{t} = \sqrt{\overline{\alpha}^{t}}\bm{z}^0 + \sqrt{1-\overline{\alpha}^{t}}\bm{\epsilon}$, where $\bm{\epsilon} \sim \mathcal{N}(\bm{0}, \bm{I})$ and $\overline{\alpha}^{t} = \textstyle{\prod^{t}_{m=1}\alpha^m}$.
In the denoising process, each denoising step is parameterized as $p_{\theta}(\bm{z}^{t-1}|\bm{z}^{t}) = \mathcal{N}(\bm{z}^{t-1}; \bm{\mu}_\theta(\bm{z}^{t},t),\bm{\Sigma}_\theta(\bm{z}^t,t))$,
where $\theta$ contains learnable parameters.

\begin{figure}[t]
\centering
\includegraphics[width=.81\columnwidth]{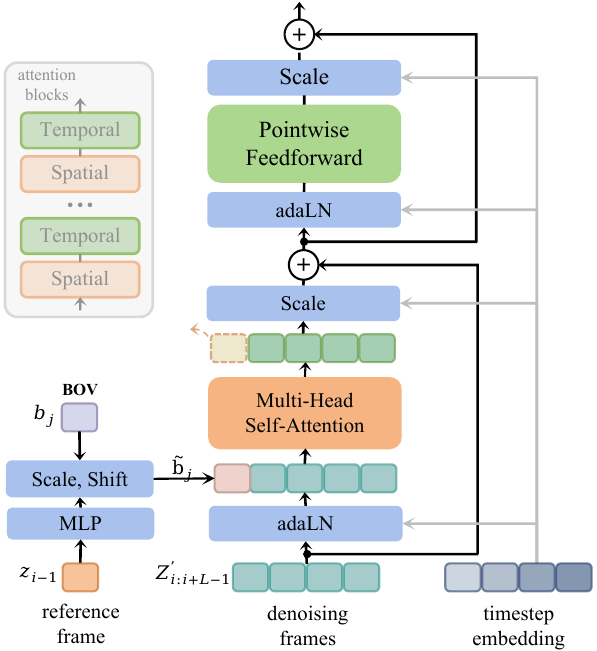}
\caption{Temporal attention block ($j$-th) with our ada-BOV token design. The BOV token contains global features and is modulated with the latest generated frame.}
\label{fig2}
\end{figure}
%

\vspace{4pt}
\noindent\textbf{Introducing Ada-BOV Tokens.}
In conventional autoregressive VDMs (either the chunk-based extension or stream denoising), the denoising process is confined to the attention window and lacks global context\footnote{Although some prior works incorporate a global reference frame~\cite{long2024videostudio, weng2024art}, they overlook the dynamic scene changes/transitions crucial for long video generation.}.
Inspired by special tokens in language models, we introduce adaptive Begin-of-Video (ada-BOV) tokens into temporal attention blocks.
\par As illustrated in Fig.~\ref{fig2}, the design incorporates two key features. First, to capture the dynamic variations across different scenes, we employ a modulation strategy. It injects a periodically updated reference frame---the latest denoised output at the end of each iteration---into the BOV tokens, using an adaLN-like mechanism.
Second, motivated by previous research showing that attention blocks focus on distinct levels of abstraction~\cite{dosovitskiy2020image, raghu2021vision}, we assign a distinct BOV token to each temporal attention block. This allows for specialized contextual representation, as each token learns to model the global information pertinent to its specific level of abstraction and extracts distinct guidance features. Combined with the modulation strategy, it achieves more coherent and effective guidance.

\begin{figure}
    \centering
    \includegraphics[width=1\linewidth]{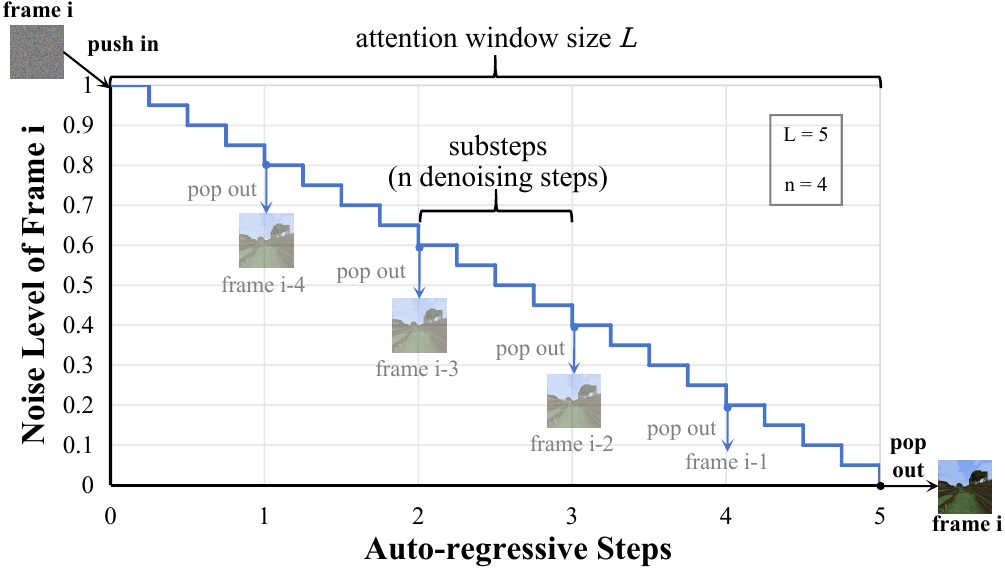}
    \vspace{-3ex}
    \caption{Illustration of the sampling trajectory for frame $i$ with our refined inference strategy. $L$ represents the attention window size, and each autoregressive iteration consists of $n$ substeps.}
    \label{fig3}
\end{figure}

\par Specifically, the unique BOV token for each ($j$-th) temporal attention block is denoted as $\bm{b}_j$. Let $\bm{Z}^{\prime}_{i:i+L-1}=[\bm{z}^{\prime}_i,\ldots \bm{z}^{\prime}_{i+L-1}]$ be the input of the $j$-th temporal attention block. The reference frame $\bm{z}_{i-1}$ is processed by a block-specific $\text{MLP}_j$ to generate modulation parameters, \ie, $(\bm{\gamma}_j, \bm{\beta}_j) = \text{MLP}_j(\bm{z}_{i-1})$. Then, $\bm{b}_j$ is modulated as 
\begin{align}
    \tilde{\bm{b}}_j = \bm{\gamma}_j \odot \bm{b}_j + \bm{\beta}_j,
\end{align}
where $\odot$ represents the Hadamard product.
Through this modulation, the local dynamics in $\bm{z}_{i-1}$ have been elegantly integrated with the global features in $\bm{b}_j$.
 Subsequently, $\tilde{\bm{b}}_j$ is prepended to the block's input sequence, denoted as $[\tilde{\bm{b}}_j, \bm{Z}^{\prime}_{i:i+L-1}]$. Then, the full attention mechanism operates on a token sequence of length $L+1$, and the output sequence excludes its first token to maintain the original length for subsequent attention blocks.
\par This design also enhances motion dynamics. It combines the informative BOV token $\bm{b}_j$ with high-fidelity guidance from clean reference $\bm{z}_{i-1}$, overcoming the low motion dynamics commonly induced by noisy references.

\subsection{Refinement Strategy for Stream Denoising}\label{sec:RSD}
We first introduce conventional stream denoising and then present our refinement strategy for stream denoising. This strategy provides high-quality local dynamics, ensuring the efficacy of our ada-BOV design (Sec.~\ref{sec:bot}).

\vspace{4pt}
\noindent\textbf{Stream Denoising.}
Let $\bm{Z}_{0:L}=[\bm{z}_0,\ldots \bm{z}_{L}]$ be given as the condition frames, where $L$ is the attention window size and constrained by the training length. $\bm{Z}_{0:L}$ is first applied progressive noise, denoted as $\bm{Z}_{0:L}^{0:L} = [\bm{z}_0,\bm{z}_1^1,\ldots,\bm{z}_{L}^T]$, where $T=L$.
Then, in the denoising process, we take $\bm{Z}_{1:L}^{1:T}$ as the initial input sequence and $\bm{z}_0$ as the initial reference frame. Correspondingly, iteration $k$ begins with an input sequence $\bm{Z}_{i:i+L-1}^{1:T}$ and a reference frame $\bm{z}_{i-1}$, where $i=k$.
In vanilla stream denoising, $\bm{Z}_{i:i+L-1}^{1:T}$ is transformed into $\bm{Z}_{i:i+L-1}^{0:T-1}$, conditioned on the reference frame $\bm{z}_{i-1}$.
This denoising process can be efficiently realized using DDIM \cite{song2021denoising} on a sparse subsequence of the total $T$ timesteps.
The subsequence $\bm{\tau}$ is typically constructed from $L+1$ uniformly distributed points spanning the interval $[0, T]$.
Then, the input sequence can be denoted as $\bm{Z}_{i:i+L-1}^{\bm{\tau}_{1:L}}$ and in the sampling step, all frames' noise levels are reduced with $\Delta t=\frac{T}{L}$, yielding $\bm{Z}_{i:i+L-1}^{\bm{\tau}_{0:L-1}}$.
By popping out $\bm{z}_i$ and appending a new pure-noise frame $\bm{z}_{i+L}$, the input sequence for the next iteration, $\bm{Z}_{i+1:i+L}^{\bm{\tau}_{1:L}}$ is constructed.

\vspace{4pt}
\noindent\textbf{Refined Inference Strategy.}
The length of the sampling trajectory (\ie, the number of denoising steps) of vanilla stream denoising is constrained to the attention window size $L$. When $L$ is small for efficiency, this short trajectory is often insufficient to produce high-quality results. 
To achieve the high-generation quality required to fully reveal the advantages of our ada-BOV design without sacrificing computational efficiency, we propose a refined inference strategy.
Fig.~\ref{fig3} visualizes the sampling trajectory with our strategy.
It decomposes the denoising process within each autoregressive iteration into $n$ sequential substeps:
\begin{align}\label{eq:rsd_transition}
    1\text{-st substep:}\quad & p_\theta(\bm{Z}_{i:i+L-1}^{\bm{\tau}_{1:L}-\Delta t^\prime} |\bm{z}_{i-1},\bm{Z}_{i:i+L-1}^{\bm{\tau}_{1:L}}), \\
    \bm{\ldots} \notag \\
    n\text{-th substep:}\quad  & p_\theta(\bm{Z}_{i:i+L-1}^{\bm{\tau}_{0:L-1}} |\bm{z}_{i-1},\bm{Z}_{i:i+L-1}^{\bm{\tau}_{0:L-1}+\Delta t^\prime}),\label{eq:nsubstep}
\end{align}
where each substep applies a uniform noise decrement of $\Delta t' = \frac{T}{n \cdot L}$, and the first frame in $\bm{Z}_{i:i+L-1}^{\bm{\tau}_{0:L-1}}$ is fully denoised by sampling $\bm{z}_i^{\tau_0}$ (\ie, $\bm{z}_i^0$) from Eq.~(\ref{eq:nsubstep}). This strategy effectively extends the sampling trajectory to $\{T, \frac{(n \cdot L-1)T}{n \cdot L}, \dots, \frac{T}{n \cdot L}\}$. It enables a more stable denoising process and significantly improves the sampling quality. Consequently, as these high-quality generated frames are used to modulate BOV tokens in subsequent iterations, the efficacy of our ada-BOV design is fully supported. Refer to Appendix~\ref{B RSD} for the formalized algorithms.

It is worth noting that using the refined inference strategy is substantially more efficient than directly extending the attention window. The time complexity of self-attention is $O(L^2)$. As the sample trajectory length is tied to the attention window size, naively extending the sample trajectory to nL would result in the complexity of $O(n^2L^2)$. Conversely, the refined inference strategy extends the sampling trajectory length to $nL$ by performing $n$ sub-steps in each iteration, without changing the attention window size, \ie, it has a complexity of $O(nL^2)$. The optimal choice for $n$ is analyzed in further detail in Sec.~\ref{sec:ablation}.

\vspace{4pt}
\noindent\textbf{Training Strategy.}
To support the refined inference strategy proposed previously, we explore a suitable training strategy.
We first present the training objectives and then investigate three noise schedules.
%
\par Current diffusion models~\cite{ho2020denoising, peebles2023scalable} are trained with the variational lower bound~\cite{kingma2013auto} of the log likelihood of $\bm{z}^0$, formulated as $\mathcal{L}(\theta) = -\log(p_\theta(\bm{z}^0|\bm{z}^1)) + \sum\nolimits_t \mathcal{D}_{KL}(q(\bm{z}^{t-1}|\bm{z}^t,\bm{z}^0)||p_\theta(\bm{z}^{t-1}|\bm{z}^t))$,
where $\mathcal{D}_{KL}$ is determined by the mean $\bm{\mu}_\theta$ and covariance $\bm{\Sigma}_\theta$. By reparameterizing $\bm{\mu}_\theta$ as a noise prediction network $\bm{\epsilon}_\theta$ and fixing $\bm{\Sigma}_\theta$ as a constant variance schedule~\cite{ho2020denoising}, the objective function can be simplified as $\mathcal{L}_{\text{simple}}(\theta)= \mathop{\mathbb{E}}_{\bm{z},\bm{\epsilon},t}\left[ \|\bm{\epsilon}_\theta(\bm{z}^t,t) - \bm{\epsilon}\|_2^2\right]$.
$\bm{\epsilon}_\theta$ is the noise predicted by the model with parameters $\bm{\theta}$.
In our setting, the input frames are partially noised with the noise levels $\bm{\tau}$. Consequently, the simplified objective function for our model is 
\begin{equation}\label{eq:oursLsimple}
    \mathcal{\tilde{L}}_{\text{simple}}(\theta)= \mathop{\mathbb{E}}_{\bm{Z},\bm{\epsilon},\bm{\tau}}\left[ \|\bm{\epsilon}_\theta(\bm{Z}^{\bm{\tau}_{0:L}}_{0:L},\bm{\tau}_{0:L}) - \bm{\epsilon}\|_2^2\right],
\end{equation}
where $\bm{\epsilon} \sim \mathcal{N}(\bm{0}, \bm{I})$. Following previous works~\cite{nichol2021improved, peebles2023scalable}, in practice, we train the model with learnable covariance $\bm{\Sigma}_\theta$ by optimizing a
combination of $\mathcal{\tilde{L}}_{\text{simple}}$ and $\mathcal{L}$.

Given the training objectives, we investigate three candidate training noise schedules (\ie, the subsequence $\bm{\tau}$ mentioned in Sec.~\ref{sec:RSD}), all of which adhere to the training-inference alignment prerequisite.
\par \textit{(1) Progressive Schedule.}
It employs a deterministic, linear noise level sequence during training. This schedule is designed to align the noise levels applied in training directly with those used during inference, ensuring a structural alignment between these two phases. Consequently, it can be represented as:
\begin{align} \label{eq:tau}
    \bm{\tau}_{1:L}=[\frac{T}{L},\dots,\frac{(L-1)T}{L},T].
\end{align}

\par \textit{(2) Random Schedule.}
Following Diffusion Forcing~\cite{chen2024diffusion}, this schedule utilizes a totally stochastic noise level sequence. During each iteration, the noise levels are randomly and independently sampled from the interval $(0, T]$:
\begin{align}
    \hat{\bm{\tau}}_{1:L}=[\hat{\tau}_1, \hat{\tau}_2, \dots, \hat{\tau}_L],
\end{align}
where $\hat{\tau}_i\sim U(1, T)$ for $i=1, 2, \dots, L$.
This stochastic schedule inherently covers a wide range of noise levels, including those encountered during inference.

\par \textit{(3) Disturbance-Augmented Schedule.}
Based on a linear noise level sequence, this schedule further applies a stochastic disturbance at each step to increase the diversity of the training samples. Following the formulation in PA-VDM~\cite{xie2025progressive}, we define the noise levels as:
\begin{align}
    \bm{\tau}^\prime_{1:L}=\bm{\tau}_{1:L}+0.4\epsilon^\prime\cdot \frac{T}{L},
\end{align}
where $\bm{\tau}$ is represented in Eq.~(\ref{eq:tau}), and $\epsilon^\prime \sim \mathcal{N}(0,1)$.
Its underlying linear structure prevents irrelevant noise level sequences, guaranteeing stable convergence. Meanwhile, incorporated perturbations fortify the model's resilience to minor prediction errors.
\begin{table*}
  \centering
  \caption{Quantitative comparison on Minecraft~\cite{guss2019minerl}. 
  We report FVD\cite{unterthiner2019fvd} and VBench\cite{huang2023vbench} scores, evaluated on 2084 videos (each containing 289 frames) generated by each model from the same test set. The best values within each group are shown in \textbf{bold}.}
  \vspace{-0.5ex}
  \begin{tabular}{lcccccccc}
    \toprule
    Method & FVD$\downarrow$ & \begin{tabular}[c]{@{}c@{}}Object\\ Consistency$\uparrow$\end{tabular} & \begin{tabular}[c]{@{}c@{}}Background\\ Consistency$\uparrow$\end{tabular} & \begin{tabular}[c]{@{}c@{}}Motion\\ Smoothness$\uparrow$\end{tabular} & \begin{tabular}[c]{@{}c@{}}Dynamic\\ Degree$\uparrow$\end{tabular} & \begin{tabular}[c]{@{}c@{}}Aesthetic\\ Quality$\uparrow$\end{tabular} & \begin{tabular}[c]{@{}c@{}}Imaging\\ Quality$\uparrow$\end{tabular} & FPS $\uparrow$ \\
    \midrule
    VDT~\cite{lu2024vdt} & 1247.75 & 0.6724 & 0.9203 & 0.9623 & \textbf{1.0} & 0.3422 & 0.3967 & 9.51 \\
    OpenSora~\cite{opensora} & 385.18 & 0.8460 & 0.9476 & 0.9767 & 0.9893 & 0.4102 & 0.4201 & 7.81 \\
    FIFO~\cite{kim2024fifo} & \textbf{261.96} & 0.8445 & 0.9498 & 0.9785 & 0.9909 & \textbf{0.4316} & \textbf{0.4269} & 7.48 \\
    Ours & 436.74 & \textbf{0.8609} & \textbf{0.9512} & \textbf{0.9816} & 0.9453 & 0.4250 & 0.4146 & \textbf{12.71} \\
    \bottomrule
  \end{tabular}
  \label{tab1}
\end{table*}
\begin{figure*}[t]
\centering
\includegraphics[width=1.\textwidth]{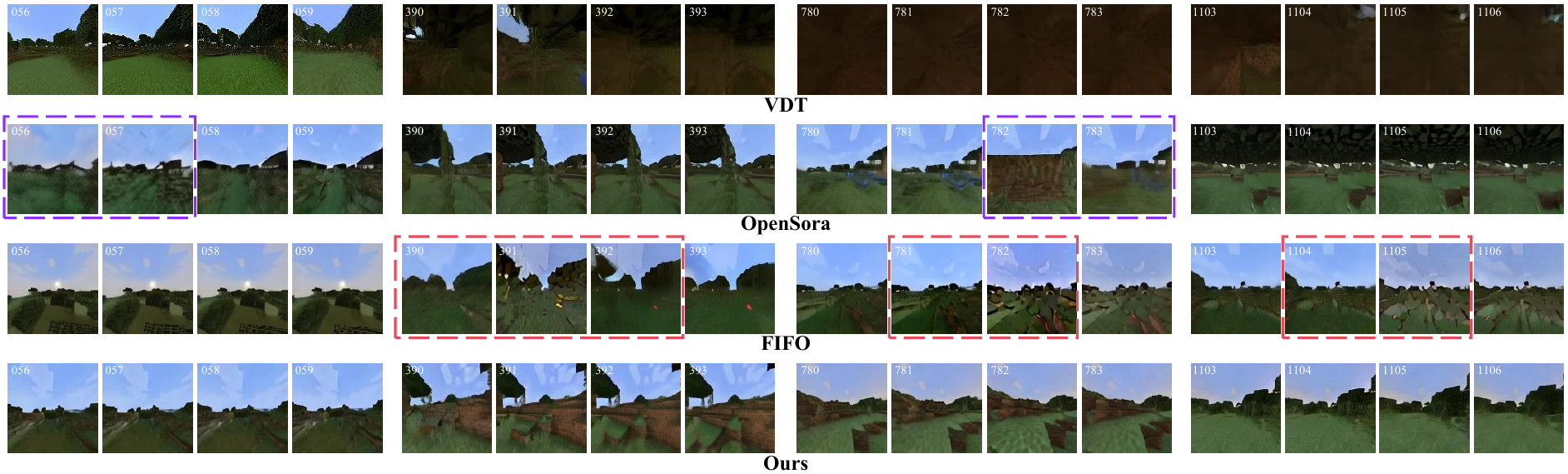} 
\vspace{-4ex}
\caption{Qualitative examples on Minecraft~\cite{guss2019minerl}, generated by VDT~\cite{lu2024vdt}, OpenSora~\cite{opensora}, FIFO~\cite{kim2024fifo}, and ours. 
We use purple boxes to represent the low-quality frames in OpenSora and red boxes to represent the flickers between adjacent frames in FIFO. 
}
\label{fig4}
\end{figure*}

\section{Experiments}
\label{sec:exp}
\subsection{Experimental Settings and Details}
\noindent\textbf{Implementation Details.}
Our model was built on diffusion models~\cite{sohl2015deep,ho2020denoising} with Transformer architecture~\cite{peebles2023scalable, ma2025latte}, and was initialized with weights from OpenSora v1.2~\cite{opensora}. 
The attention window size $L$ was set to 16. The maximum number of diffusion timesteps is $T=1000$.
Notably, since our work focused on unconditional video generation, the text encoder was excluded from the model architecture. The number of substeps in every iteration was $n=4$, and it was investigated in ablation studies in Sec.~\ref{sec:ablation}. More details are provided in Appendix~\ref{sec:ma}.

\vspace{4pt}
\noindent\textbf{Dataset.}
Our model was evaluated on Minecraft~\cite{guss2019minerl} and Sky Timelapse~\cite{zhang2020dtvnet}, two challenging benchmarks widely adopted for evaluating long video generation and prediction models. 
Minecraft provides high diversity, viewpoint changes, and complex motion dynamics. Meanwhile, Sky Timelapse provides high-resolution, dynamic, and real-world scenes. Based on the above properties and considering the available computational resources, we did not train on open-domain videos. More details about dataset preprocessing are provided in Appendix~\ref{sec:data}.


\vspace{4pt}
\noindent\textbf{Evaluation Metrics.}
We evaluated the overall generation quality using Fréchet Video Distance (FVD)~\cite{unterthiner2019fvd} from a distributional perspective. To conduct a more fine-grained analysis, we selected six key dimensions from VBench~\cite{huang2023vbench}: For temporal consistency, we evaluated Object Consistency, Background Consistency, Motion Smoothness, and Dynamic Degree. For spatial quality, we used Aesthetic Quality and Imaging Quality. Additionally, we conducted a systematic user study as a supplementary human-centric metric, calculating preference rates based on three criteria.

\subsection{Comparisons with State-of-the-Arts}
We first compared the generation quality of our method against six baselines, namely VDT~\cite{lu2024vdt}, SEINE~\cite{chen2023seine}, OpenSora~\cite{opensora}, FIFO~\cite{kim2024fifo}, DiTCtrl~\cite{cai2025ditctrl}, and Ca2-VDM~\cite{gao2025ca2}.
Refer to Appendix~\ref{sec:bs} for more baseline details.

\begin{figure*}[t]
\centering
\includegraphics[width=1.\textwidth]{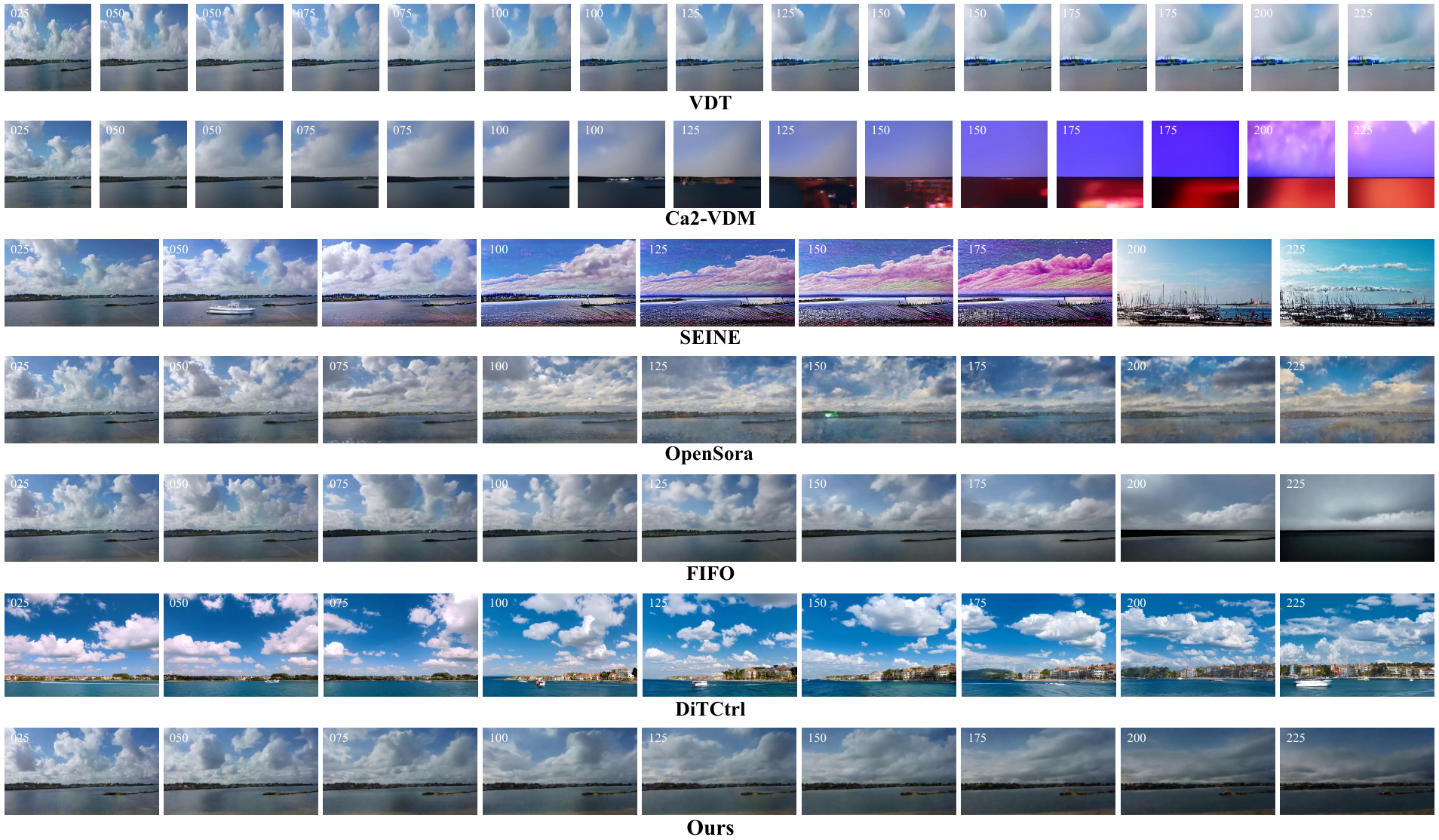} 
\vspace{-4ex}
\caption{Qualitative examples on Sky Timelapse~\cite{zhang2020dtvnet}, generated by VDT~\cite{lu2024vdt}, Ca2-VDM~\cite{gao2025ca2}, SEINE~\cite{chen2023seine}, OpenSora~\cite{opensora}, FIFO~\cite{kim2024fifo}, DiTCtrl~\cite{cai2025ditctrl} and ours. The baseline methods exhibit significant flaws, including noticeable artifacts in OpenSora, a gradual collapse into static frames in VDT, and severe quality degradation or unnatural color shift in the remaining models.}
\vspace{-1ex}
\label{fig9}
\end{figure*}

\vspace{4pt}
\noindent\textbf{Quantitative Comparison.}
As shown in Table~\ref{tab1}, our method achieves the highest scores in Motion Smoothness, Object Consistency, and Background Consistency, highlighting its superior performance in maintaining temporal and global coherence. Furthermore, it demonstrates a significant advantage in inference speed. FIFO~\cite{kim2024fifo} scores well on per-frame quality metrics. However, its qualitative results are unsatisfactory, particularly in terms of temporal coherence (\cf~Fig.~\ref{fig4} and Fig.~\ref{fig9}).

\vspace{4pt}
\noindent\textbf{Qualitative Comparison.}
We use a sequence depicting wandering around the terraced fields and a time-lapse of a complex seaside town as the initial conditions to highlight the performance differences. The results reveal that VDT's~\cite{lu2024vdt} outputs suffer from a gradual collapse into static frames. And videos generated by Ca2-VDM~\cite{gao2025ca2} and SEINE~\cite{chen2023seine} exhibit rapid temporal collapse.
Although OpenSora~\cite{opensora} maintains better stability, it still produces low-fidelity content (\cf~purple boxes in Fig.~\ref{fig4}) and visible artifacts (\cf~frame 100$\sim$225 in Fig.~\ref{fig9}).
While FIFO~\cite{kim2024fifo} achieves high per-frame quality, its susceptibility to error accumulation leads to flickering (\cf~red boxes in Fig.~\ref{fig4}) and noticeable temporal degradation (\cf~frame 125$\sim$225 in Fig.~\ref{fig9}).
DiTCtrl~\cite{cai2025ditctrl}, benefiting from additional text prompts, delivers strong imaging quality but suffers from noticeable unnatural color shift over time (\cf~frame 25$\sim$100 in Fig.~\ref{fig9}). Moreover, its computational inefficiency severely limits its practicality.
In contrast, our method successfully preserves both high frame quality and temporal stability, demonstrating its significantly better ability in maintaining long-term coherence and global consistency. It also achieves a significant speed-up over DiTCtrl.
\begin{figure}[t]
\centering
\includegraphics[width=1.\columnwidth]{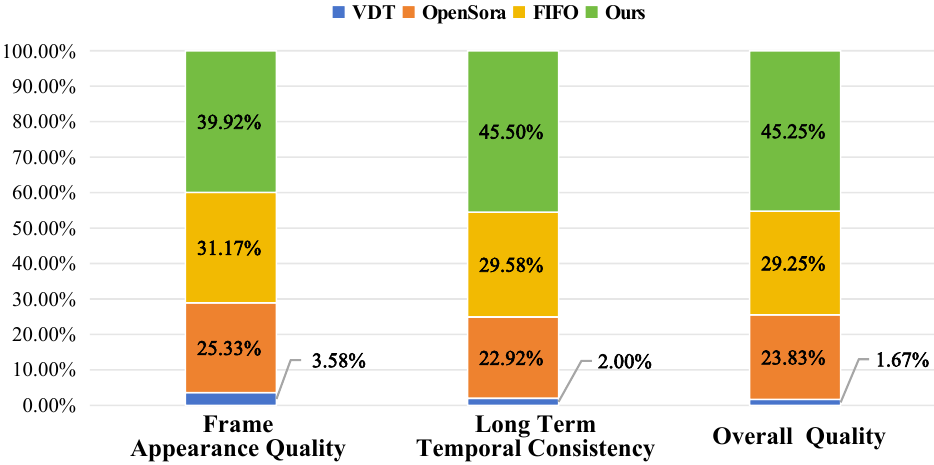}
\vspace{-3.5ex}
\caption{User study results. Our method achieves higher preference rates than the others across all criteria.}
\label{fig4-2}
\end{figure}

\vspace{4pt}
\noindent\textbf{User Study.}
We further conducted a comprehensive user study to assess which model produced the most appealing and coherent videos from a human-centric perspective.
Specifically, we recruited 40 participants and asked them to evaluate 30 sets of videos according to two criteria: 1) \textbf{Frame Appearance Quality}, and 2) \textbf{Long Term Temporal Consistency}. 
We select VDT~\cite{lu2024vdt}, OpenSora~\cite{opensora}, and FIFO~\cite{kim2024fifo}, the top three baselines according to their superior qualitative results and inference efficiency, as our competitors.
Each set comprises four videos generated by our method and three competitors under the same conditions. To ensure fairness, the videos within each set were randomly shuffled. Participants were first asked to select the best video for each criterion, and subsequently choose the one that demonstrated the best \textbf{Overall Quality}, considering both criteria. 
As summarized in Fig.~\ref{fig4-2}, our method is consistently preferred by participants, achieving the highest preference rates across all three criteria. 
For example, our model outperforms FIFO in Long Term Temporal Consistency  (45.50\% \vs 29.58\%) and OpenSora in Overall Quality (45.25\% \vs 23.83\%).
This demonstrates that it shows a clear advantage in generating both visually appealing and temporally coherent videos.

\subsection{Ablation Studies} \label{sec:ablation}
\begin{figure}[t]
\centering
\includegraphics[width=1.\columnwidth]{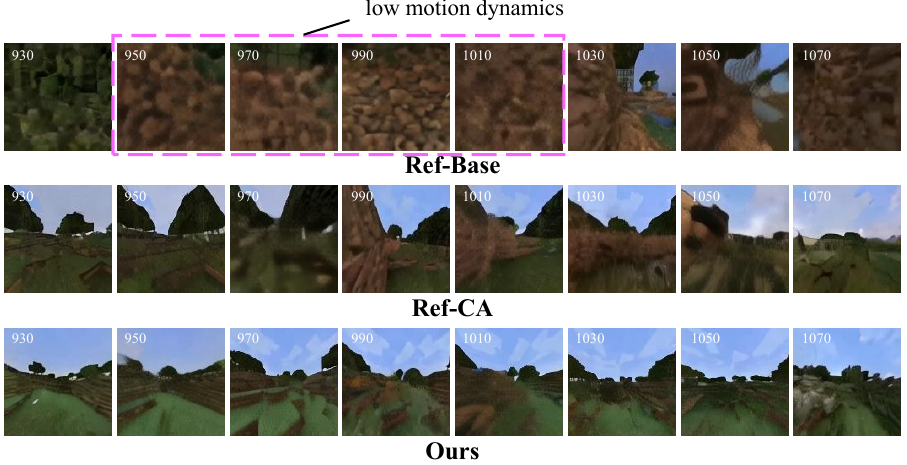} 
\vspace{-4ex}
\caption{Qualitative examples generated by Ref-Base, Ref-CA, and our method. We use a pink box to represent the video clip with low motion dynamics in Ref-Base. The corresponding quantitative results are presented in the first three rows of Table~\ref{tab2}.}
\label{fig5}
\end{figure}

\noindent\textbf{Ada-BOV Token Design.}
To evaluate its effectiveness, we designed two baselines with different reference injection styles. 1) \textbf{Ref-Base}: It was constructed upon OpenSora and was able to denoise video sequences with independent per-frame noise levels. No additional reference was used. 2)\textbf{Ref-CA}: Built upon Ref-Base, it injected additional global guidance into every attention block via cross-attention, simulating conventional methods for maintaining global consistency. And the first frame was used as its fixed global reference.
The comparison results in Table~\ref{tab2} demonstrate our method's superior performance, as it achieves the best FVD score and outperforms the Ref-CA variant across all VBench dimensions. We note that our model scores lower than Ref-Base on certain metrics like Background Consistency. 
We argue that this stems from the Ref-Base model's extremely low score on Dynamic Degree. It suppresses inter-frame differences, consequently leading to misleading high scores on consistency metrics that are sensitive to such differences. 
Qualitative results in Fig.~\ref{fig5} clearly show Ref-Base's low motion dynamics (\cf~the pink box), and ours performs substantially better than it.
In terms of efficiency, our method achieves comparable inference speed with Ref-Base (12.71 \vs 12.66). And it is 20.7\% faster than Ref-CA. The whole efficiency results are provided in Appendix~\ref{sec:er} (\cf~Table~\ref{tab:er}).


\begin{table*}
  \centering
  \caption{Quantitative results on Minecraft~\cite{guss2019minerl}. The first three rows show the comparison between ours and two baselines. The last three rows present an ablation study of our training noise schedules. The best values within each group are shown in \textbf{bold}, and the second-best values within each group are shown with \underline{underline}. Ref-Base performs poorly in Dynamic Degree, which leads to misleadingly high scores on other metrics (\cf~Sec.~\ref{sec:ablation}). The Progressive schedule scores well, but contains multiple flickers (\cf~Fig.~\ref{fig7}).}
  \vspace{-0.5ex}
  \begin{tabular}{l|ccccccccc}
    \toprule
    Method & \begin{tabular}[c]{@{}c@{}}Training\\ Noise Schedule\end{tabular} & FVD$\downarrow$ & \begin{tabular}[c]{@{}c@{}}Object\\ Consistency$\uparrow$\end{tabular} & \begin{tabular}[c]{@{}c@{}}Background\\ Consistency$\uparrow$\end{tabular} & \begin{tabular}[c]{@{}c@{}}Motion\\ Smoothness$\uparrow$\end{tabular} & \begin{tabular}[c]{@{}c@{}}Dynamic\\ Degree$\uparrow$\end{tabular} & \begin{tabular}[c]{@{}c@{}}Aesthetic\\ Quality$\uparrow$\end{tabular} & \begin{tabular}[c]{@{}c@{}}Imaging\\ Quality$\uparrow$\end{tabular} \\
    \midrule
    Ref-Base & Rand. & 782.20 & \textbf{0.8723} & \textbf{0.9537} & \textbf{0.9832} & 0.8244 & 0.4080 & \underline{0.4214} \\
    Ref-CA & Rand. & 611.51 & 0.8352 & 0.9435 & 0.9781 & \underline{0.9705} & 0.3828 & 0.3644 \\
    \midrule
    \multirow{3}{*}{Ours} & Rand. & 446.86 & 0.8439 & 0.9479 & 0.9776 & \textbf{0.9823} & 0.4008 & 0.4023 \\ 
    & Progre. & \underline{445.64} & \underline{0.8677} & \underline{0.9530} & \underline{0.9822} & 0.9341 & \textbf{0.4326} & \textbf{0.4259} \\
    & Dist. Aug. & \textbf{436.74} & 0.8609 & 0.9512 & 0.9816 & 0.9453 & \underline{0.4250} & 0.4146 \\
    \bottomrule
  \end{tabular}
  \label{tab2}
\end{table*}

\begin{figure*}[t]
\centering
\includegraphics[width=1.\textwidth]{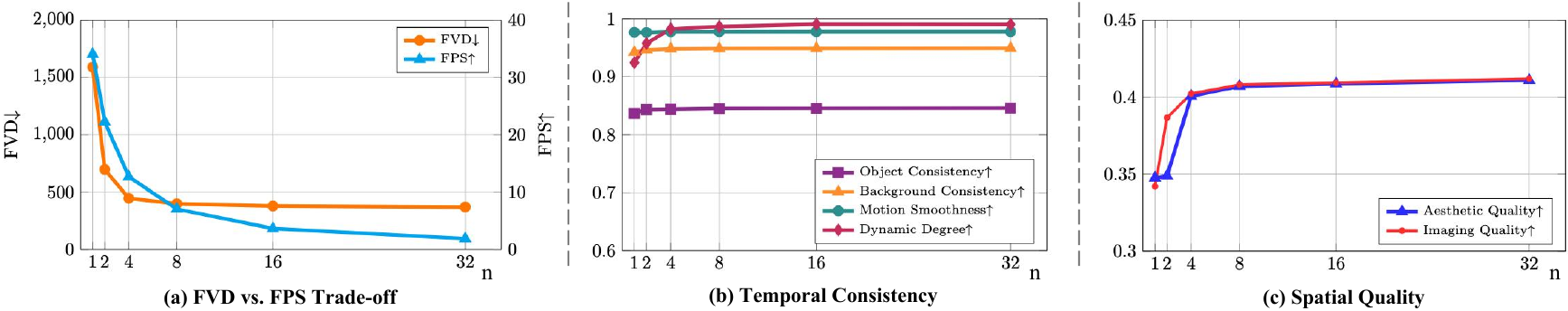}
\caption{FVD~\cite{unterthiner2019fvd}, FPS, and VBench metrics~\cite{huang2023vbench} w.r.t. parameter $n$. The case where $n=1$ corresponds to vanilla stream denoising. And the results for $n>1$ demonstrate the effectiveness of the refined inference strategy. Chart (a) clearly illustrates the trade-off between quality improvement and computational cost.}
\vspace{-1.5ex}
\label{fig6}
\end{figure*}
\begin{figure}[t]
\centering
\includegraphics[width=1.\columnwidth]{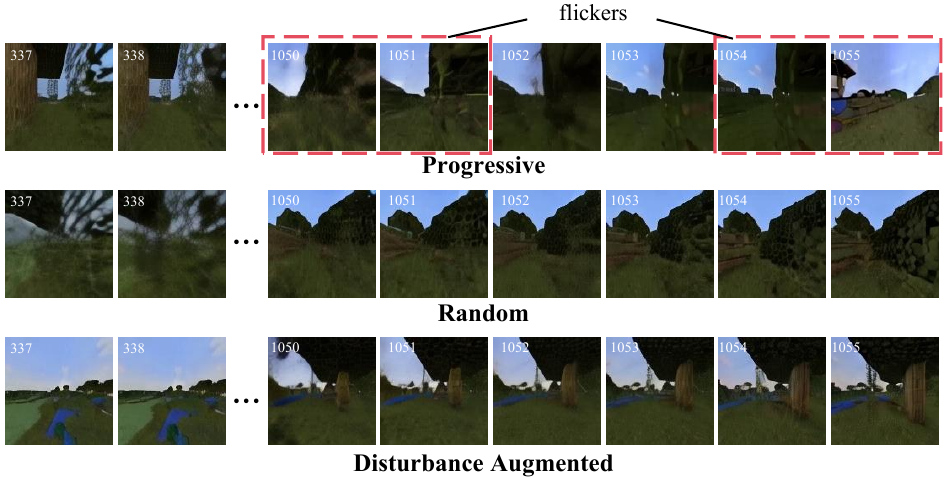} %
\vspace{-3.5ex}
\caption{Qualitative examples with Progressive, Random, and Disturbance-Augmented schedule. Flickers appear frequently in the results of the Progressive schedule. The corresponding quantitative results are presented in the last three rows of Table~\ref{tab2}. }
\label{fig7}
\end{figure}

\vspace{4pt}
\noindent\textbf{Refined Inference Strategy.}
We evaluated the effect of the number of substeps ($n$) on both generation quality and computational cost. In particular, $n=1$ represented the vanilla stream denoising. 
The results in Fig.~\ref{fig6} reveal a classic trade-off between generation quality and efficiency. Specifically, as $n$ grows, the generation quality improves while the inference speed decreases. 
Notably, the quality gains diminish significantly when $n>4$, while the computational cost (\ie, the inference speed) continues to increase linearly.
Therefore, we choose $n=4$ as the optimal balance between performance and computational cost. 
More results are provided in Appendix~\ref{sec:rs} (\cf~Table~\ref{tab:ab} and Fig.~\ref{fig8}).

\vspace{4pt}
\noindent\textbf{Training Noise Schedule.}
\label{sec:dat}
We compared the three training noise schedules introduced in Sec.~\ref{sec:RSD}. The results are shown in Fig.~\ref{fig7} and the last three rows of Table~\ref{tab2}.
The Progressive schedule effectively streamlines training convergence, but its inherent lack of noise diversity leaves the model vulnerable to subsequent error accumulation, resulting in noticeable flickers in the generated videos (\cf~red boxes in Fig.~\ref{fig7}).
The alternative Random schedule inherently provides high diversity, leading to a more robust model that does not exhibit flicker. Nevertheless, it severely impedes convergence and incurs prohibitive computational overhead.
The Disturbance-Augmented schedule effectively balances these trade-offs. The model trained with it achieves faster convergence while producing flicker-free videos. It also quantitatively outperforms the Random schedule across nearly all evaluation metrics.
\section{Conclusions}
\label{sec:con}
In this paper, we propose a novel approach for autoregressive long video generation. We utilize ada-BOV tokens for flexible and robust global guidance, achieving better temporal coherence and motion dynamics. Meanwhile, we propose a refinement strategy for stream denoising to enhance visual fidelity and reveal the advantages of our ada-BOV design. Our approach achieves state-of-the-art results, producing high-quality and coherent videos. This work also provides valuable insights into training noise schedules for the community of streaming video generation.
\vspace{4pt}
\par \noindent \textbf{Limitations.} Despite its promising results, our work has two primary limitations. Firstly, the conditioning on video alone restricts real-time control, which could be enhanced by incorporating multimodal inputs, such as text prompts or keyboard signals. Secondly, the ability to generate more realistic and diverse environments remains a key area for exploration towards building a general-purpose world model.
\clearpage
{
    \small
    \bibliographystyle{ieeenat_fullname}
    \bibliography{main}
}

\clearpage
\begin{appendices}
\section{Experimental Details}    
\subsection{Model and Architecture} \label{sec:ma}
Our model was built on diffusion models~\cite{sohl2015deep,ho2020denoising} with Transformer architecture~\cite{peebles2023scalable, ma2025latte}, and initialized with weights from OpenSora v1.2~\cite{opensora}. We utilized the 2D VAE from LDM~\cite{rombach2022high} and the 3D VAE from MAGVIT-v2~\cite{yu2024language}. The attention window size $L$ was set to 16 with a maximum timestep of $T=1000$. Notably, since our work focuses on the task of unconditional video generation, the text encoder was excluded from the model architecture. 
We adopted the majority of the default training configurations from the pre-trained model, including the AdamW optimizer~\cite{loshchilov2018decoupled} and its corresponding warm-up schedule. For efficiency, we employed a mixed-precision training strategy~\cite{micikevicius2018mixed} with the bfloat16 format. For reproducibility, the random seed was set to 42 across all experiments. The remaining model hyperparameters are presented in detail in Table~\ref{tab:hyper}.

\begin{table}[H] \small
    \centering
    \caption{Hyperparameters of our video diffusion model.}
    \scalebox{1.0}{
    \begin{tabularx}{1.0\linewidth}{ p{0.5\linewidth} | X<{\centering} }
    \toprule
    \multicolumn{2}{c}{Hyperparameters} \\
    \midrule
    Hidden Size             & 1152  \\
    Patch Size              & (1,2,2)  \\
    Depth                   & 28    \\
    Num Heads               & 16    \\
    $\beta$ Linear Start    & 0.0001    \\
    $\beta$ Linear End      & 0.02    \\
    Number of Diffusion Timesteps           & 1000  \\
    \bottomrule
    \end{tabularx}
    }
    \label{tab:hyper}
\end{table}

\subsection{Experimental Environments} \label{sec:ee}
All experiments were conducted on a system equipped with a single NVIDIA A6000 GPU (48GB) and an Intel Xeon Platinum 8352V CPU. The software environment consisted of Ubuntu 22.04 LTS (Linux Kernel: 6.8.0-52-generic), PyTorch 2.2.2, and CUDA 12.1.66.

\subsection{Dataset Details and Data Preprocessing} \label{sec:data}
We conducted our experiments on the public datasets Minecraft~\cite{guss2019minerl} and Sky Timelapse~\cite{zhang2020dtvnet}, two challenging benchmarks widely adopted for evaluating long video generation and prediction models.
Minecraft consists of over 194,000 high-quality videos, each with a duration of 301 frames and a spatial resolution of $128 \times 128$ pixels. This dataset is characterized by its significant diversity, substantial viewpoint changes, and complex motion dynamics. Such characters make it a challenging benchmark for evaluating model performance in controlled but complex environments.
Sky Timelapse contains 2,392 real-world videos with a high resolution of $360 \times 640$ pixels. They provide dynamic, real-world scenes with natural motion, serving as a testbed for evaluating the model's generalization capability on high-resolution, real-world data.

Our data preprocessing filtered videos to meet the minimum length requirement imposed by the models' architecture. Specifically, to accommodate positional variations within the 3D VAE~\cite{yu2024language}, the input latent sequence must be at least twice the length of the attention window (16). Consequently, we discarded all videos in the training set shorter than 119 frames, calculated as $\lceil16\times2\div5\rceil\times 17=119$, where $\lceil\cdot\rceil$ is the ceiling operation. For the test set, this threshold was increased to $\lceil16\times4\div5\rceil\times 17=221$ frames for FIFO~\cite{kim2024fifo}, which required a larger context, while the threshold for all other evaluations remained at 119 frames.

\subsection{Baseline Settings} \label{sec:bs}
We detail the configurations for all baseline models in this section. All models generate videos of resolution 128$\times$128 on Minecraft, and videos of resolution 360$\times$640 on Sky Timelapse unless otherwise specified.
\begin{itemize}
    \item  VDT~\cite{lu2024vdt}. It natively supports Sky Timelapse~\cite{zhang2020dtvnet}, and we finetuned its official checkpoint on Minecraft~\cite{guss2019minerl}. The condition length was set to 8, and the resolution on Sky Timelapse was set to 256$\times$256.
    \item OpenSora~\cite{opensora}. It provides general-purpose checkpoints. Since we excluded its text encoder, we finetuned it on both datasets with a condition length of 8.
    \item SEINE~\cite{chen2023seine}. It originally sets the condition length to 1, and we expanded the condition length to 8.
    \item FIFO~\cite{kim2024fifo}. We implemented this training-free method on the OpenSora v1.2~\cite{opensora} backbone, and set $n$ to 4.
    \item Ca2-VDM~\cite{gao2025ca2}. We directly employed its official checkpoint and prescribed settings on Sky Timelapse, without modification. The results' resolution was 256$\times$256.
    \item DiTCtrl~\cite{cai2025ditctrl}. It provides general-purpose checkpoints. We used the same conditional frames from Sky Timelapse and generated corresponding text prompts with GPT-4o~\cite{hurst2024gpt}. The results' resolution was 540$\times$720.
\end{itemize}



\begin{algorithm*}[t]
\caption{Inference procedure of the vanilla stream denoising paradigm}\label{alg:1}
    \begin{algorithmic}[1]
    
    \REQUIRE Initial latent frames $\displaystyle \bm{Z}_{0:L} = \{\bm{z}_0, \bm{z}_1, ..., \bm{z}_L\}$ with attention window size $L$, and the maximum noise level $T$. Random Noise $\displaystyle \bm{\epsilon} \sim \mathcal{N}( \mathbf{0} , \bm{I} )$.
    \item[] 
    
    \STATE $\displaystyle \bm{\tau}_{0:L} = \left\{\bm{\tau}_0, \bm{\tau}_1, \ldots, \bm{\tau}_L \right\} = \left\{ 0, \frac{T}{L}, \ldots, T\right\}$ 
    \hfill \COMMENT{\# Linear sampling noise level schedule~~$\qquad\qquad~\quad$}
    
    
    \STATE $\displaystyle \bm{Z}^{\bm{\tau}_{1:L}}_{1:L} = \sqrt{\alpha^{\bm{\tau}_{1:L}}}\bm{Z}_{1:L} + \sqrt{1-\alpha^{\bm{\tau}_{1:L}}}\bm{\epsilon}$
    \hfill \COMMENT{\# Add noise and set to progressive noise levels~~$\quad~\quad$}

    \item[] 
    \FOR[\hfill \COMMENT{\# Every iteration decreases the noise level by $1\cdot \frac{T}{L}~~~$}]{each generatio iteration $\displaystyle k = 1, 2, \ldots, K$}
        \STATE $\displaystyle p_\theta\left(\bm{Z}^{\bm{\tau}_{0:L-1}}_{k:k+L-1} \middle| \bm{z}_{k-1}, \bm{Z}^{\bm{\tau}_{1:L}}_{k:k+L-1} \right)$
        
        \STATE $\displaystyle \bm{z}_{k+L}^T \sim \mathcal{N}( \mathbf{0} , \bm{I} )$
        \hfill \COMMENT{\# Sample a new noisy frame~~$\qquad\qquad\qquad\qquad~~\quad$}
        
        \STATE Append $\bm{z}_k$ to the list of clean frames
        
        \STATE $\displaystyle \bm{Z}^{\bm{\tau}_{1:L}}_{k+1:k+L} = \left\{ \bm{z}_{k+1}^{\bm{\tau}_1}, \ldots, \bm{z}_{k+L-1}^{\bm{\tau}_{L-1}}, \bm{z}_{k+L}^{T} \right\}$ \hfill \COMMENT{\# Remove $\bm{z}_{k-1}$, shift forward, and append $\bm{z}_{k+L}^T~~~\quad$}
        
    \ENDFOR
    \item[] 
    \RETURN List of $K$ clean frames $\{\bm{z}_1, \bm{z}_2,\cdots, \bm{z}_K\}$.

    \end{algorithmic}
\end{algorithm*}


\vspace{-1ex}
\begin{algorithm*}[t]
\caption{Inference procedure of the refinement strategy for stream denoising}\label{alg:2}
    \begin{algorithmic}[1]
    
    \REQUIRE Initial latent frames $\displaystyle \bm{Z}_{0:L} = \{\bm{z}_0, \bm{z}_1, ..., \bm{z}_L\}$ with the attention window size $L$, and maximum noise level $T$. Random noise $\displaystyle \bm{\epsilon} \sim \mathcal{N}( \mathbf{0} , \bm{I} )$. $\Delta t^\prime=\frac{T}{n\cdot L}$.
    \item[] 
    
    \STATE $\displaystyle \bm{\tau}_{0:L} = \left\{\bm{\tau}_0, \bm{\tau}_1, \ldots, \bm{\tau}_L \right\} = \left\{ 0, \frac{T}{L}, \ldots, T\right\}$
    
    
    \STATE $\displaystyle \bm{Z}^{\bm{\tau}_{1:L}}_{1:L} = \sqrt{\alpha^{\bm{\tau}_{1:L}}}\bm{Z}_{1:L} + \sqrt{1-\alpha^{\bm{\tau}_{1:L}}}\bm{\epsilon}$ 
    \item[] 
    \FOR[\hfill \# Every iteration decreases the noise levels by $n\cdot \frac{T}{n\cdot L}$]{each generation iteration $\displaystyle k = 1, 2, \ldots, K$}
        \FOR{each sub-denoising step $\displaystyle i = 1, 2, \ldots, n$}
        
        \STATE $\displaystyle p_\theta\left(\bm{Z}^{\bm{\tau}_{1:L}-i\cdot \Delta t^{\prime}}_{k:k+L-1} \middle| \bm{z}_{k-1}, \bm{Z}^{\bm{\tau}_{1:L}-(i-1)\cdot \Delta t^{\prime}}_{k:k+L-1} \right)$ \hfill \COMMENT{\# Every substep decreases the noise levels by $\Delta t^{\prime}$\qquad}

        \ENDFOR
        \item[] 
        
        \STATE $\displaystyle \bm{z}_{k+L}^T \sim \mathcal{N}( \mathbf{0} , \bm{I} )$ 
        
        \STATE Append $\bm{z}_k$ to the list of clean frames
        
        \STATE $\displaystyle \bm{Z}^{\bm{\tau}_{1:L}}_{k+1:k+L} = \left\{ \bm{z}_{k+1}^{\bm{\tau}_1}, \ldots, \bm{z}_{k+L-1}^{\bm{\tau}_{L-1}}, \bm{z}_{k+L}^{T} \right\}$
        
    \ENDFOR

    \item[] 
    \RETURN List of $K$ clean frames $\{\bm{z}_1, \bm{z}_2,\cdots, \bm{z}_K\}$.

    \end{algorithmic}
\end{algorithm*}

\subsection{Evaluation Settings} \label{sec:es}
\par For our qualitative evaluation, the models were trained on the entire training set (194,051 videos). During inference, every model generated 2,084 videos (289 frames) conditioned on the entire test set. FVD~\cite{unterthiner2019fvd} and VBench~\cite{huang2023vbench} were computed on these videos to ensure a comprehensive and robust evaluation.
\par For the user study, 30 videos were randomly selected from the test set to serve as conditions. Each model was then tasked with generating a corresponding set of 30 videos, each consisting of 1,190 frames.

\section{Algorithms of Stream Denoising} \label{B RSD}
This section illustrates pseudo-code for vanilla stream denoising (Algorithm~\ref{alg:1}) and stream denoising with our refined inference strategy (Algorithm~\ref{alg:2}).

\section{Additional Results}
\subsection{Efficiency Results} \label{sec:er}
\par To evaluate computational efficiency, we measured the total time required to generate 30 videos of 289 frames, denoted as $\overline{T}$. FPS was subsequently calculated as: $\text{FPS}=(30\times 289)/\overline{T}$. The performance of the different models is presented in Table~\ref{tab:er}.

\subsection{Examples for the Refinement Strategy} \label{sec:rs}
\begin{figure*}[t]
\centering
\includegraphics[width=1.\textwidth]{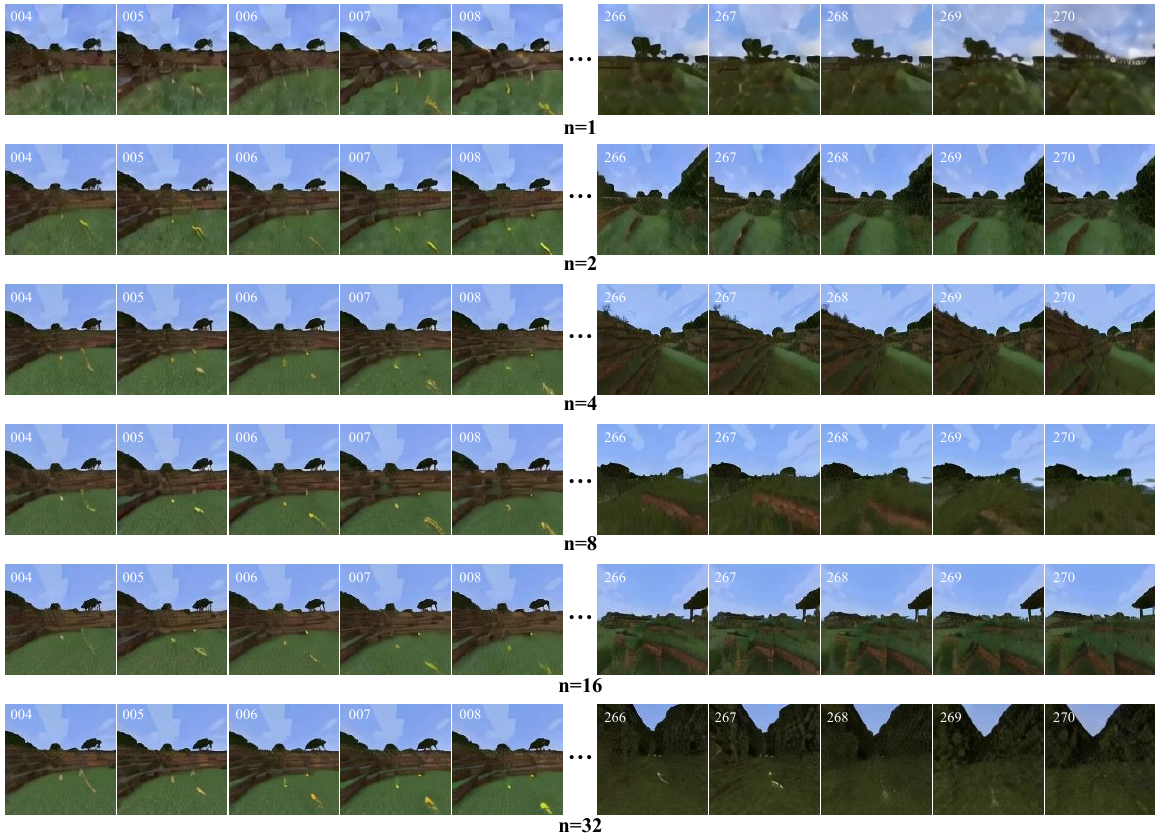}
\caption{Qualitative examples for different substeps $n$ on Minecraft~\cite{guss2019minerl}.}
\label{fig8}
\end{figure*}

This section complements results of the ablation study on the number of substeps ($n$) in Section~\ref{sec:ablation}. We provide the detailed quantitative metrics in Table~\ref{tab:ab} and corresponding qualitative examples in Fig.~\ref{fig8}, which were omitted from the main text for brevity. Additional video examples for all experiments are available in the supplementary material.

\vspace{-1ex}
\begin{table}[H] \small
    \centering
    \caption{Efficiency results for different models on Minecraft~\cite{guss2019minerl}. We measured the total time required to generate 30 videos of 289 frames, and subsequently calculated their FPS. }
    \scalebox{1.0}{
    \begin{tabularx}{1.0\linewidth}{ p{0.5\linewidth} | X<{\centering} }
    \toprule
    \multicolumn{2}{c}{Inference Speed (FPS)$\uparrow$} \\
    \midrule
    VDT~\cite{lu2024vdt}             & 9.51  \\
    OpenSora~\cite{opensora}              & 7.81  \\
    FIFO~\cite{kim2024fifo}                   & 7.48    \\
    Ref-Base               & 12.66    \\
    Ref-CA    & 10.53    \\
    Ours      & 12.71    \\
    \bottomrule
    \end{tabularx}
    }
    \label{tab:er}
\end{table}

\begin{table*}[t]
  \centering
  \begin{tabular}{ccccccccc}
    \toprule
    $n$ & FVD$\downarrow$ & \begin{tabular}[c]{@{}c@{}}Object\\ Consistency$\uparrow$\end{tabular} & \begin{tabular}[c]{@{}c@{}}Background\\ Consistency$\uparrow$\end{tabular} & \begin{tabular}[c]{@{}c@{}}Motion\\ Smoothness$\uparrow$\end{tabular} & \begin{tabular}[c]{@{}c@{}}Dynamic\\ Degree$\uparrow$\end{tabular} & \begin{tabular}[c]{@{}c@{}}Aesthetic\\ Quality$\uparrow$\end{tabular} & \begin{tabular}[c]{@{}c@{}}Imaging\\ Quality$\uparrow$\end{tabular} & \begin{tabular}[c]{@{}c@{}}Inference\\ Speed(FPS)$\uparrow$\end{tabular} \\
    \midrule
    1 & 1590.70 & 0.8365 & 0.9422 & 0.9697 & 0.9243 & 0.3476 & 0.3421 & 34.07 \\
    2 & 699.22 & 0.8433 & 0.9461 & 0.9764 & 0.9574 & 0.3849 & 0.3867 & 22.29 \\
    4 & 446.86 & 0.8439 & 0.9479 & 0.9776 & 0.9823 & 0.4008 & 0.4023 & 12.71 \\
    8 & 398.61 & 0.8452 & 0.9486 & 0.9777 & 0.9865 & 0.4071 & 0.4081 & 7.07 \\
    16 & 379.87 & 0.8454 & 0.9488 & 0.9778 & 0.9906 & 0.4088 & 0.4094 & 3.68 \\
    32 & 370.77 & 0.8458 & 0.9491 & 0.9778 & 0.9901 & 0.4110 & 0.4120 & 1.91 \\
    \bottomrule
  \end{tabular}
  \caption{Quantitative results for different substeps $n$ on Minecraft~\cite{guss2019minerl}.}
  \label{tab:ab}
\end{table*}

\end{appendices}

\end{document}